\documentclass[wcp]{jmlr}

\usepackage{longtable}

\usepackage{booktabs}

\usepackage{lineno}

\makeatletter
\let\Ginclude@graphics\@org@Ginclude@graphics 
\makeatother

\jmlrvolume{222}
\jmlryear{2023}
\jmlrworkshop{ACML 2023}

\title[Short Title]{TLMCM Network for Medical Image Hierarchical Multi-Label Classification}

 \author{\Name{Meng Wu$^*$} \Email{mwu344@gatech.edu}\\
  \Name{Siyan Luo$^*$} \Email{sluo96@gatech.edu}\\
  \Name{Qiyu Wu$^*$} \Email{qwu346@gatech.edu}\\
  \addr College of Computing Georgia Tech
  \AND
  \Name{Wenbin Ouyang} \Email{wenbinoy@gmail.com}\\
  \addr Redmond WA, US 
  }

\editors{Berrin Yan{\i}ko\u{g}lu and Wray Buntine}

\begin{document}

\maketitle
\def\thefootnote{*}\footnotetext{These authors contributed equally to this work.}

\begin{abstract}
    Medical Image Hierarchical Multi-Label Classification (MI-HMC) is of paramount importance in modern healthcare, presenting two significant challenges: \textit{data imbalance} and \textit{hierarchy constraint}. Existing solutions involve complex model architecture design or domain-specific preprocessing, demanding considerable expertise or effort in implementation. To address these limitations, this paper proposes Transfer Learning with Maximum Constraint Module (TLMCM) network for the MI-HMC task. The TLMCM network offers a novel approach to overcome the aforementioned challenges, outperforming existing methods based on the $AU\overline{(PRC)}$(Area Under the Average Precision and Recall Curve) metric. In addition, this research proposes two novel accuracy metrics, $EMR$(Exact Match Ratio) and $HammingAccuracy$, which have not been extensively explored in the context of the MI-HMC task. Experimental results demonstrate that the TLMCM network achieves high multi-label prediction accuracy($80\%$-$90\%$) for MI-HMC tasks, making it a valuable contribution to healthcare domain applications. 
\end{abstract}

\section{Introduction}

Hierarchical Multi-label Classification (HMC) is a classification task that involves hierarchically organized classes. In the domain of healthcare, the Medical Image Hierarchical Multi-label Classification (MI-HMC) is important for efficient image interpretation, retrieval, and diagnosis \cite{Cai01,Kim01}. The MI-HMC problem naturally arises in the medical industry and academia, given that X-ray images \cite{Chen01}, and microscope images \cite{Dimitrovski01} can incorporate tree-structured sub-categories. However, MI-HMC faces two key challenges: \textit{data imbalance} and  \textit{hierarchy constraint} \cite{Giunchiglia01}. Existing solutions involve complex model architectures \cite{Wehrmann01,Noor01} or domain-specific preprocessing \cite{Dimitrovski02,Quan01,pelka18}. 

In prior research, the emphasis has predominantly leaned towards generic solutions, often overlooking the specific intricacies of MI-HMC tasks. In our study, we introduce a novel approach, the Transfer Learning with Maximum Constraint Module (TLMCM) network, which squarely tackles the challenges inherent to the MI-HMC domain.

The TLMCM network combines a pretrained deep learning CNN model with a Maximum Constraint Module (MCM) as proposed by \cite{Giunchiglia01}. It effectively addresses the issue of data imbalance by harnessing the power of transfer learning techniques, which have previously demonstrated their efficacy on small image datasets. The MCM method we employ is meticulously designed to ensure the satisfaction of the "hierarchy constraint" in multi-label prediction results, and it boasts a straightforward implementation. One of the key advantages of the TLMCM network is that it obviates the need for extensive image preprocessing or domain-specific knowledge for feature extraction prior to model training.

For generic HMC tasks, Area Under Precision-Recall Curve ($AU\overline{(PRC)}$) is the typical evaluation metric \cite{Giunchiglia01,Wehrmann01}. In specific MI-HMC tasks, where each prediction follows a distinct path in a hierarchical label structure, we introduce two new accuracy metrics: $EMR$ and $HammingAccuracy$ \cite{Sorower01}, for a comprehensive evaluation. We thoroughly assessed the TLMCM network using these three metrics on two MI-HMC tasks with X-ray image datasets (ImageCLEF09A and ImageCLEF09D) \cite{clef}. Our experiments demonstrate the superior performance of the TLMCM network compared to the current state-of-the-art methods \cite{Giunchiglia01}. Moreover, it achieves exceptionally high accuracy in multi-label predictions across both tasks, highlighting the practical significance of this research.


The key contributions of this work are: 1) the proposal of the compact and highly effective TLMCM network, which adeptly addresses common MI-HMC challenges and outperforms state-of-the-art methods in our experimental tasks; 2) the introduction of two novel evaluation metrics, $EMR$ and $HammingAccuracy$, facilitating intuitive accuracy assessment in the MI-HMC domain, an area where such metrics have been largely unexplored. 
\section{Related Work}
Current Hierarchical Multi-Label Classification (HMC) methods use local, global, or hybrid approaches. Local methods build separate classifiers for each node, while global methods use one classifier for the full hierarchy. Hybrids combine both \cite{pmlr-v80-wehrmann18a}. Recent work shows directly incorporating hierarchical information in model improves performance \cite{Giunchiglia01}. 
 For medical images, Hierarchical Medical Image Classification uses stacked deep learning models \cite{Kowsari01}, and deep Hierarchical Multi-Label Classification targets chest x-ray diagnosis \cite{Chen01}. Other work focuses on specific diseases \cite{Gour01} or body parts \cite{Hou01}.

\section{Methodology}

 We integrate the Maximum Constraint Module(MCM) as proposed in \cite{Giunchiglia01}, with the transfer learning process of a pretrained Convolutionnal Neural Network model ResNet50 \cite{He02} as the backbone. We make careful and essential adaptations to the architecture of ResNet50, for the purpose of addressing the unique challenges posed by the MI-HMC task, and further augmenting the model's capabilities in handling hierarchical multi-label classification.

\subsection{MCM Method}\label{sec:mcm}

A generic HMC task 
 must respect the \textit{hierarchy constraint}, i.e. for each label that is predicted to be true, all the ancestor labels as pre-defined in the hierarchy structure must also be predicted to be true \cite{Giunchiglia01}. 
In this regard, we adopted the two key concepts proposed in  \cite{Giunchiglia01}: \textit{Max Constraint Module (MCM)} and  \textit{Maximum Constraint Loss (MCLoss)}.

Formally, for a generic HMC task with a set $S$ of $n$ labels in total, given a label $A$, let $D_A$ be the set of all labels which are the descendants of the label $A$ in the hierarchical structure, and a machine learning model $h$ predicts the label $A$ to be true with the probability of $h_A$, we then impose the MCM module on top of the output of the model $h$, such that the output of MCM for label $A$ is:
\begin{equation}\label{eq:mcm}
\operatorname{MCM}_A=\max_{B \in \mathcal{D}_A}\left(h_B\right)
\end{equation}

In addition, we can also guide the training process by incorporating the MCM constraint into the loss function of the underlying model $h$. Formally, let $y_A$ and $y_B$
be the ground truth value of label $A$ and $B$, the loss for label A is

\begin{equation}\label{eq:loss_a}
\begin{split}
\operatorname{MCLoss}_A = &-y_A \ln \left(\max _{B \in \mathcal{D}_A}\left(y_B h_B\right)\right) \\
&-\left(1-y_A\right) \ln \left(1-\mathbf{M C M}_A\right)
\end{split}
\end{equation}

Then the final loss function is defined as 

\begin{equation}\label{eq:loss_all}
\mathrm{MCLoss}=\sum_{A \in \mathcal{S}} \mathrm{MCLoss}_A
\end{equation}

According to  \cite{Giunchiglia01}, the novel $\mathrm{MCLoss}$ function could bring benefits to the gradient backpropagation to achieve a lower loss than the standard binary cross-entropy loss function.

\subsection{Transfer Learning}\label{sec:transferlearning}

For the backbone model, we retain the ResNet50 convolution blocks unchanged and focus on adapting the last fully-connected linear layer. 
In particular, we replace the original last layer with two fully-connected linear layers with a ReLU activation in between and a Sigmoid after. The hidden dimension is set to 256, and the output dimension is equal to the total number of labels. The Sigmoid layer converts the output scores into probabilities, as required by the MCM module.

\begin{figure*}
    \begin{center}
    \fbox{\rule{0pt}{0.5in} \includegraphics[width=14cm]{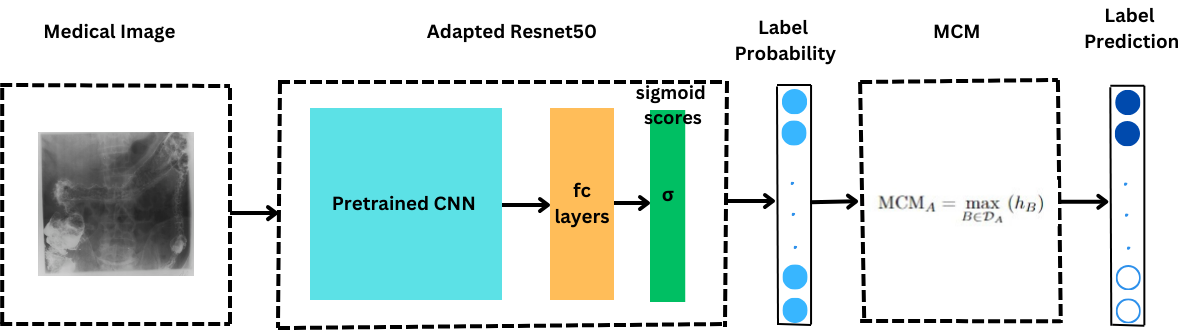} \rule{.08\linewidth}{0pt}}
    \end{center}
       \caption{TLMCM network architecture. The predicted labels are represented as dark blue solid circles, while the other non-predicted labels are shown as hollow circles.}
    \label{fig:proposed_method}
\end{figure*}

The modified ResNet50 output is then fed into the MCM to predict the probability of truth for each label. We use the adapted loss function, $\mathrm{MCLoss}$, as defined in Equation (\ref{eq:loss_all}).
The output are the probabilities for each label to be true. 
In the specific task of MI-HMC, since the label structure is \textit{mono-hierarchical}(fully explained in the dataset description section \ref{sec:datasets}), we 
just make our prediction that all labels corresponding to the \textit{maximum} probability of the MCM output are predicted to be true. 
The whole architecture design is shown in Figure \ref{fig:proposed_method}.

There are two common approaches to transfer learning: (1)freezing the convolution module as the fixed feature extractor, and only training the linear classifier(ResNet50 as Fix Feature Extractor, hereafter referred to as RNFFE); (2) fine-tuning the convolution module with the pre-trained weights of the convolution module as well as the linear classifier module(ResNet50 Fine-Tuning, hereafter referred to as RNFT).
In the experiment section \ref{sec:expSetup}, we will explore whether RNFFE or RNFT is more suitable for the MI-HMC task.

\section{Experiments}

\subsection{Datasets}\label{sec:datasets}

The main dataset that we used for this research is the 2009 ImageCLEF edition of the IRMA X-ray dataset \cite{clef}. 
Each image in the dataset was classified based on the IRMA code \cite{Lehmann01}. 
A classification code may consist of three or four digits, representing a \textit{mono-hierarchical} classification structure for the corresponding medical image. 
A \textit{mono-hierarchical} label means that the classification hierarchy is a tree structure, thus each child node can only have one parent node.  


In this study, our focus was directed toward evaluating the effectiveness of our model on two specific classification codes: anatomical (A) of 110 labels, and directional (D) of 36 labels. The dataset consists of 14410 images in total, and we split it into the train/validation/test set with the ratio of 70:15:15.
This dataset choice of the (A) and (D) codes also allowed for a direct comparison with the baseline results\footnote{The ImageCLEF07 datasets from their research are an older version no longer available. However, the ImageCLEF09 datasets share the same chracteristics.}. 



\subsection{Evaluation Metrics}\label{sec:metric}


Area Under the average Precision and Recall Curve($AU\overline{(PRC)}$)  \cite{Giunchiglia01,Sorower01} is most commonly used in multi-label classification and tasks, 
and is also the metric that we used to compare with the state-of-the-art baseline results.
Additionally, Exact Match Ratio($EMR$) \cite{Sorower01} is a strict accuracy metric in that the prediction is considered correct only when the set of labels of prediction exactly matches the corresponding set of labels of ground truth. Formally, the $EMR$ is computed as follows:

\begin{equation}\label{eq:emr}
EMR = \frac{1}{n} \sum_{i=1}^n I\left(Y_i=Z_i\right)
\end{equation}
where, $I$ is the indicator function, $n$ is total number of all labels, $Y_i$ is the set of ground true labels for sample $i$, and $Z_i$ is the set of predicted labels for sample $i$.

From the application perspective of the MI-HMC task, if a prediction is partially correct, it still deserves some credit. 
In this respect, we utilize the multi-label classification $HammingAccuracy$ \cite{Sorower01}, which is defined as the following with the same meaning as all notations defined in Eq. \ref{eq:emr}:

\begin{equation}\label{eq:hamming}
HammingAccuracy=\frac{1}{n} \sum_{i=1}^n \frac{\left|Y_i \cap Z_i\right|}{\left|Y_i \cup Z_i\right|}
\end{equation}



\subsection{Experiment setup}\label{sec:expSetup}

We performed experiments with both RNFFE and RNFT on the MI-HMC tasks of ImageCLEF09A and ImageCLEF09D. We maintained the same learning rate of $5e-6$, weight decay of $1e-6$, batch size of 32 and the total number of epoches of 120 for each task. The ReLU activation was used for the linear classifier and no dropout was applied.
Then we trained the model with the Adam optimizer. 

The training was completed on the Google Cloud Platform(GCP) virtual machine. We used the regular configuration of the ``n1-standard-4" machine type, and 1 NVIDIA T4 GPU.  We finished 4 training sessions(2 tasks, 2 approaches) in 10 hours, which is reasonably computationally efficient. The code is at \url{https://github.com/flowing-time/IMAGE-HMC}.

\subsection{Results and discussions}
The $AU\overline{(PRC)}$, $EMR$, and $hammingAccuracy$ for the tasks of ImageCLEF09A and ImageCLEF09D are summarized in Table \ref{fig:test_result}.
On both tasks, the transfer learning approach RNFT shows a significantly higher $AU\overline{(PRC)}$ score than the baseline results of C-HMCNN(h) in  \cite{Giunchiglia01} and some other recent models with good performance(e.g. \cite{Wehrmann01}, \cite{pelka18}). 

\begin{table*}
    \begin{tabular}{c|c|ccccc}
    \hline
    Task & Metric & \shortstack[c]{RNFFE} & \shortstack[c]{RNFT} & \shortstack[c]{Giunchiglia\\(2020)} & \shortstack[c]{Wehrmann\\(2018)} & \shortstack[c]{Pelka\\(2018)}\\ \hline
    ~ & $AU\overline{(PRC)}$ & 0.858  & \textbf{0.984}   & 0.956 & 0.950 & N/A\\ 
    \shortstack[c]{ImageCLEF09A} & $EMR$ & 0.449  & \textbf{0.789}  & N/A  & N/A & 0.603\\ 
    ~ & $Hamming$ & 0.642  & \textbf{0.890}  & N/A  & N/A & N/A\\ \hline
    ~ & $AU\overline{(PRC)}$ & 0.928  & \textbf{0.984} & 0.927 & 0.920 & N/A\\ 
    \shortstack[c]{ImageCLEF09D} & $EMR$ & 0.451  & \textbf{0.880}  & N/A & N/A & 0.791\\ 
    ~ & $Hamming$ & 0.636  & \textbf{0.917}  & N/A & N/A & N/A\\ \hline
    \end{tabular}
    \caption{Test datasets result in summarization. The best results are in bold.}
    \label{fig:test_result}
\end{table*}

To the best of our knowledge, the $EMR$ and $hammingAccuracy$ metrics have not been commonly reported on the previous HMC tasks. 
Nevertheless, our TLMCM network shows 80\% - 90\% prediction accuracy on the MI-HMC task and demonstrated its great value in real-world healthcare applications.


\begin{figure*}
    \begin{center}
    \includegraphics[scale=0.3]{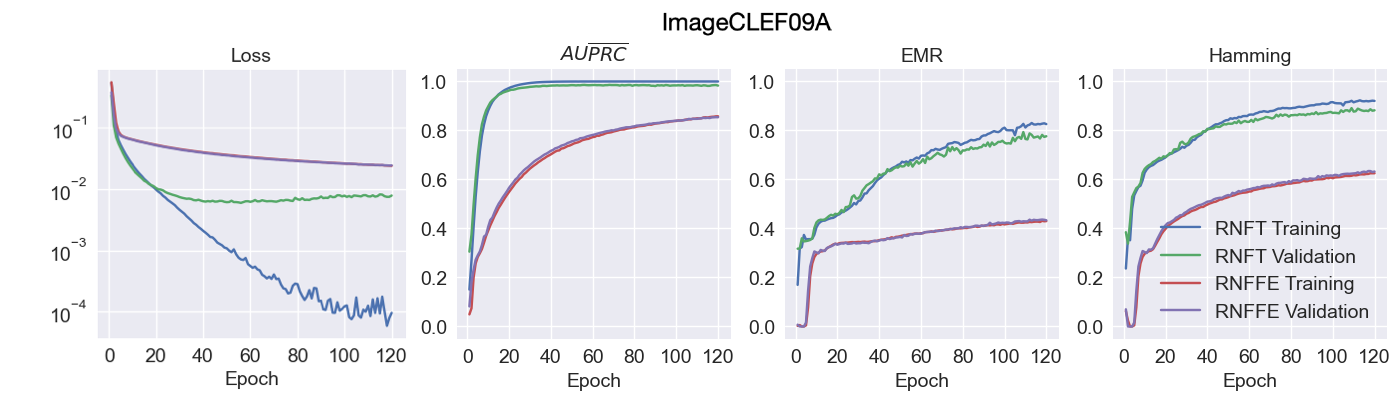}
    \end{center}
    \caption{Learning curves from from the left to right: loss, $AU\overline{(PRC)}$, $EMR$, $Hamming$}
    \label{fig:learningCurve}
\end{figure*}

For a better understanding of fine-turning approaches in our methodology, we plotted the learning curves of both approaches for the ImageCLEF09A task in Figure \ref{fig:learningCurve}. The logarithmic scale is used for the loss(1st colummn), while linear scale is used for the metrics(2nd, 3rd, 4th column).
It is clearly shown that significantly lower training and validation loss can be achieved by the RNFT approach than RNFFE. 
In the 2nd column, we can see RNFT quickly attains the plateau of the highest $AU\overline{(PRC)}$ score, for both the training and validation set. The 3rd and 4th columns show the prediction accuracy($EMR$ and $HammingAccuracy$) over the training iterations and the RNFT approach is again the winner. 

Overall, the experiments show that for the MI-HMC task, our TLMCM network is highly effective, and the transfer learning approach of fine-tuning with pretrained weights(RNFT) should be the prioritized choice, as it does not only perform best in all three metrics but also achieves the best result with the fewest training iterations.

\section{Conclusion}

We proposed a novel 
TLMCM network for hierarchical multi-label classification on radiological medical images. To the best of our knowledge, TLMCM network is the first method to incorporate the Maximum Constraint Module (MCM) approach and transfer learning in medical image classification, eliminating the requirement for domain-specific knowledge in medical image pre-training. Our architecture outperformed current state-of-the-art models on the benchmark tasks of ImageCLEF09A, ImageCLEF09D. Furthermore, we introduced the metrics of $EMR$ and $Hamming Accuracy$ for evaluating the performance of TLMCM network in the context of the MI-HMC problem. 
As part of our future work, our proposed architecture can be extended to larger medical image datasets that have deeper levels of hierarchy. Additionally, we plan to evaluate the TLMCM network on HMC image datasets in other domains such as object recognition, fashion and clothing, extending its application beyond radiological medical images to explore its potential in diverse fields.

%
%

\bibliography{acml23}

%
%
%
%

\end{document}